\pdfoutput=1
\documentclass{article} 
\usepackage{iclr2025_conference,times}
\iclrfinalcopy

\usepackage{amsmath,amsfonts,bm}

\def\eqref#1{equation~\ref{#1}}

\def\1{\bm{1}}

\DeclareMathAlphabet{\mathsfit}{\encodingdefault}{\sfdefault}{m}{sl}
\SetMathAlphabet{\mathsfit}{bold}{\encodingdefault}{\sfdefault}{bx}{n}

\newcommand{\E}{\mathbb{E}}

\usepackage{url}

\usepackage{comment}
\usepackage{graphicx}
\usepackage{enumitem}
\usepackage{subcaption}
\usepackage{booktabs}
\usepackage{tabularx}
\usepackage{caption}
\usepackage{geometry,lmodern}
\usepackage[export]{adjustbox}

\usepackage{cleveref}

\title{Sample what you can't compress}

\author{Vighnesh Birodkar\\
\texttt{vighneshbirodkar@gmail.com}
\And
Gabriel Barcik\\
Google
\And
James Lyon\\
Google
\And
Sergey Ioffe\\
xAI
\And
David Minnen\\
Google
\And
Joshua V. Dillon\\
\texttt{jvdillon@gmail.com}
}

\newcommand{\di}{D_{\mathrm{Initial}}}
\newcommand{\dr}{D_{\mathrm{Refine}}}
\newcommand{\algoname}{SWYCC}

\newcommand{\Z}{\mathcal{Z}}

\renewcommand{\E}{\operatorname{\mathsf E}}

\newcommand{\reals}{\mathbb{R}}

\newcommand\scalemath[2]{\scalebox{#1}{\mbox{\ensuremath{\displaystyle #2}}}}

\newcommand{\Uniform}{\operatorname{Uniform}}
\newcommand{\MVN}{\operatorname{MVN}}

\newcommand{\defeq}{\stackrel{\scalemath{0.5}{\mathrm{def}}}{=}}

\DeclareFixedFont{\ttb}{T1}{txtt}{bx}{n}{10} 
\DeclareFixedFont{\ttm}{T1}{txtt}{m}{n}{10}  

\usepackage{color}
\definecolor{deepblue}{rgb}{0,0,0.5}
\definecolor{deepred}{rgb}{0.6,0,0}
\definecolor{deepgreen}{rgb}{0,0.5,0}

\usepackage{listings}

\newcommand\pythonstyle{\lstset{
language=Python,
basicstyle=\ttm,
morekeywords={self},              
keywordstyle=\ttb\color{deepblue},
emph={MyClass,__init__},          
emphstyle=\ttb\color{deepred},    
stringstyle=\color{deepgreen},
frame=tb,                         
showstringspaces=false
}}

\lstnewenvironment{python}[1][]
{
\pythonstyle
\lstset{#1}
}
{}

\newcommand\pythoninline[1]{{\pythonstyle\lstinline!#1!}}

\begin{document}

\maketitle

\begin{abstract}
For learned image representations, basic autoencoders often produce blurry results. Reconstruction quality can be improved by incorporating additional penalties such as adversarial (GAN) and perceptual losses. Arguably, these approaches lack a principled interpretation. Concurrently, in generative settings diffusion has demonstrated a remarkable ability to create crisp, high quality results and has solid theoretical underpinnings (from variational inference to direct study as the Fisher Divergence). Our work combines autoencoder representation learning with diffusion and is, to our knowledge, the first to demonstrate  \textit{jointly learning
a continuous encoder and decoder under a diffusion-based loss
and showing that it can lead to higher compression and better generation.}. 
We demonstrate that this approach yields better reconstruction quality as compared to GAN-based
autoencoders while being easier to tune. 
We also show that the resulting representation is easier to model
with a latent diffusion model as compared to the representation obtained from a state-of-the-art GAN-based loss.
Since our decoder is stochastic, it can generate details not encoded in the otherwise deterministic latent representation; we therefore name our approach ``Sample what you can't compress'', or \algoname{} for short.


\end{abstract}

\section{Introduction}\label{sec:intro}

Image autoencoders ultimately necessitate a pixel-level loss to measure and minimize distortion. A common choice is to use mean squared error (MSE). This is a problem for image and video models because MSE favors low frequencies over high frequencies
\citep{mse_blog_post}. Although generalized robust loss functions have been developed \citep{BarronCVPR2019}, they are insufficient on their own for avoiding blurry reconstructions. A popular fix is to augment a pixel-level loss with additional penalties. Typically, MSE is still used because it is easy to optimize due to its linear gradient.

For example, \citet{stable_diffusion} use a combination of MSE, perceptual
loss, and adversarial loss. \citet{vqgan} noted that an adversarial loss
helps them get high-quality images with realistic textures.
Unfortunately GANs remain challenging to train; which was most
recently noted by \citet{gigagan}, when they couldn't naively scale up 
their architecture.
The diversity of their outputs is also limited, because modern GAN based decoders
are deterministic, and thus lack the capacity to sample multiple different possibilities.

\begin{minipage}{0.47\linewidth}
    \centering
     \includegraphics[width=\textwidth,valign=t]{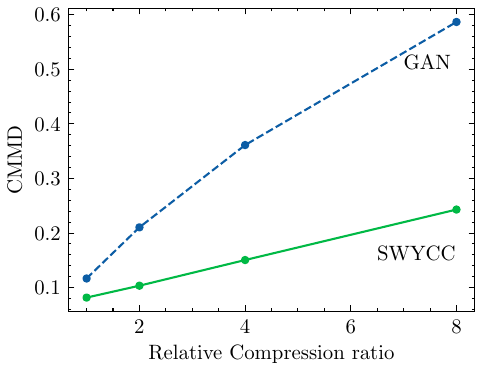}
     \captionof{figure}{Reconstruction distortion (lower is better) as a function of compression for \algoname{} and GAN based auto-encoders.}
     \label{fig:compression}
\end{minipage}
\hfill
\begin{minipage}{0.47\linewidth}
    \centering
     \includegraphics[width=\textwidth,valign=t]{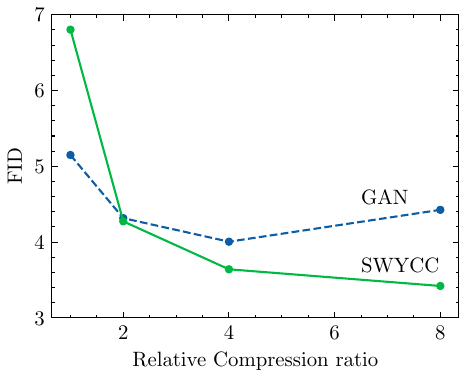}
     \captionof{figure}{Class conditional generation  quality (lower is better) as a function of compression for \algoname{} and GAN based auto-encoders.}
     \label{fig:generation}
\end{minipage}

As an alternative, this paper describes a technique for using a diffusion loss to learn an autoencoder. The diffusion loss is sensible because it is a proper scoring rule with favorable theoretical properties such as being formally connected to KL divergence \citep{SohlDickstein2015DeepUL}.
It has proven itself capable of generating crisp results with high perceptual quality as reflected by human evaluation studies \citep{simple_diffusion, diffusion_training_dynamics}.

To demonstrate its simplicity, we take a popular encoder architecture
\citet{maskgit} and marry it with a U-Net decoder, a popular denoising
architecture for diffusion \citep{simple_diffusion}.
With some additional details outlined in \cref{sec:method},
we show that this approach achieves lower distortion at all
compression levels as measured by the CMMD metric \citep{cmmd}.
Because our decoder is able to sample details at test-time that are not encoded in the latents, we call our approach ``Sample what you can't compress'' or \algoname{} for short.
\begin{figure}[h]
\centering
\begin{subfigure}{0.25\textwidth}
\includegraphics[width=\textwidth]{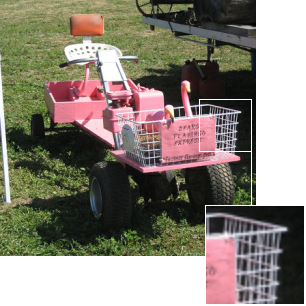}
\includegraphics[width=\textwidth]{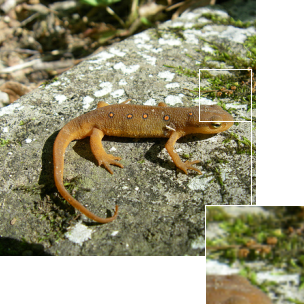}
\includegraphics[width=\textwidth]{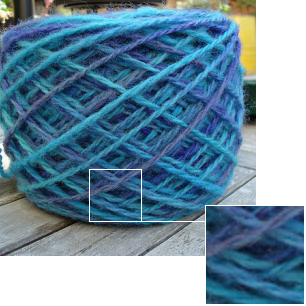}
\caption{Groundtruth}
\end{subfigure}
\begin{subfigure}{0.25\textwidth}
\includegraphics[width=\textwidth]{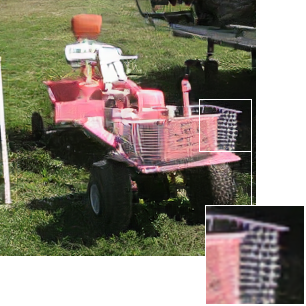}
\includegraphics[width=\textwidth]{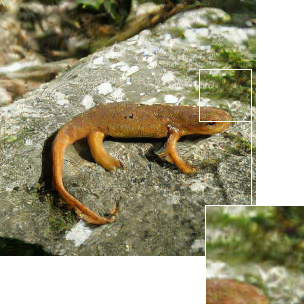}
\includegraphics[width=\textwidth]{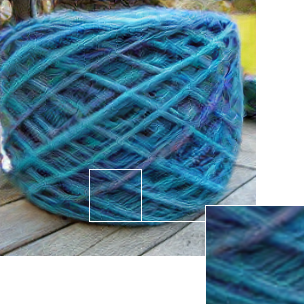}
\caption{GAN}
\end{subfigure}
\begin{subfigure}{0.25\textwidth}
\includegraphics[width=\textwidth]{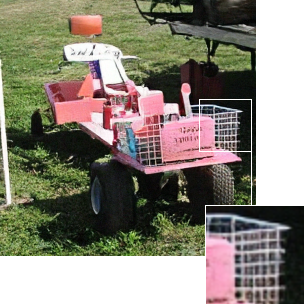}
\includegraphics[width=\textwidth]{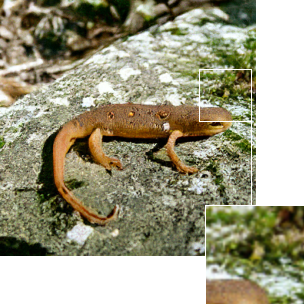}
\includegraphics[width=\textwidth]{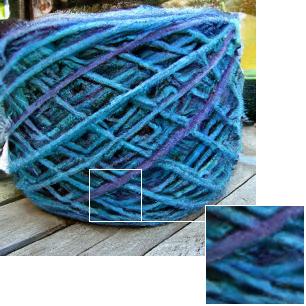}
\caption{SWYCC}
\end{subfigure}
\caption{
Comparison of GAN versus \algoname~reconstructions at $8\times$ relative compression
level. You can see that the GAN based autoencoder loses a significant amount of
detail in the highlighted portions, which \algoname{} is able to sample effectively.}
\label{fig:pictures}
\end{figure}

This work will show that,
\begin{itemize}[]
    \item SWYCC achieves lower reconstruction distortion at all tested compression levels vs SOTA GAN-based autoencoders (\cref{sec:compression}).
    \item SWYCC representations enable qualitatively better latent diffusion generation results at higher compression levels vs SOTA GAN-based autoencoders (\cref{sec:ldm}).
    \item Splitting the decoder into two parts improves training dynamics (\cref{sec:di} and \ref{sec:dr}).
\end{itemize}

\section{Method}\label{sec:method}

Eliding its various parametrizations for brevity, the standard diffusion loss is characterized by the Monte Carlo approximation of the following loss,
\begin{align}
\ell(x) \defeq \E_{\varepsilon\sim\MVN(0,I_{h\cdot w\cdot 3}), t\sim \Uniform[0,1]} \left[w_t\left\|x - D(\alpha_t x+\sigma_t\varepsilon, t)\right\|_2^2\right].
\label{eq:diffusion-loss}
\end{align}
Herein $x \in \reals^{h\times w\times 3}$ denotes a natural image and $D$ is a neural network fit using gradient descent and which serves to denoise the corrupted input $x_t\defeq \alpha_t x + \sigma_t \varepsilon$ at a given noise-level $t$.
Let the corruption process be the cosine schedule \citep{diffusion_compression_luca}, $\sigma_t^2\defeq1-\alpha_t^2$ and $\alpha_t\defeq\cos(at+b(1-t))$ where $a=\arctan(e^{10})$ and $b=\arctan(e^{-10})$.

We extend this definition to the task of autoencoding by simply allowing the denoising function to take an additional argument, $\di(E(x))$, itself having access to the uncorrupted input $x$ but only through the bottlenecking function $E$. The result is,
\begin{align}
\label{eq:aeloss}
\ell_{\mathrm{AE}}(x) \defeq \E_{\varepsilon\sim\MVN(0,I_{h\cdot w\cdot 3}), t\sim \Uniform[0,1]} \left[w_t\left\|x - \dr(\alpha_t x+\sigma_t\varepsilon, t, \di(E(x))) \right\|_2^2\right].
\end{align}

As the notation suggests $E$ is an encoder which, notably, is learned 
jointly with ``diffusion decoder'' $\dr$ and secondary decoder $\di:\Z\to\reals^{h\times w\times 3}$. The specification of $\di$ is largely a convenience but also merits secondary advantages. By mapping $z=E(x)$ back into $x$-space, we can simply concatenate the corrupted input $x_t$ and its ``pseudo reconstruction,'' $\di(z)$. Additionally, we find that directly penalizing $\di(z)$, as described below, speeds up training.

\subsection{Architectural Details}
{\bf Encoder:} We use a fully convolutional encoder in all of our
experiments, whose specifics we borrow from MaskGIT\citep{maskgit}.
The encoder consists of multiple ResNet\citep{resnet} blocks stacked on top
of each other, with GeLU\citep{gelu} for its non-linearities and GroupNorm\citep{groupnorm} for training stability.
The ResNet blocks are interspersed with strided convolutions
with stride $2$ which achieves a $2 \times$ downsampling
by itself.
To get the $8\times8$ patch size, we use 4 ResNet blocks with 3 downsampling
blocks.
The encoder architecture is common for all of our experiments,
and we only change the number of channels at the output layer
to achieve the desired compression ratio.

{\bf Decoder:} For the decoder in the GAN baseline and $\di$ we use an architecture that is 
the reverse of the encoder. For up-sampling,
we use the depth-to-space operation.
Just like the encoder, we have 
4 ResNet blocks interspersed with 
3 depth-to-space operations. 
For $\dr$ we use a U-Net as defined
by \citet{simple_diffusion}.
The U-Net has 4 ResNet blocks for downsampling
and corresponding 4 ResNet blocks for upsampling with residual connections
between blocks of the same resolution.
After 4 downsampling stages, when 
resolution is $16\times16$, we
use a self-attention block to give
the network additional capacity.

\subsection{Reducing distortion using additional distance metrics}

\begin{figure}
    \centering
    \includegraphics[width=0.9\textwidth]{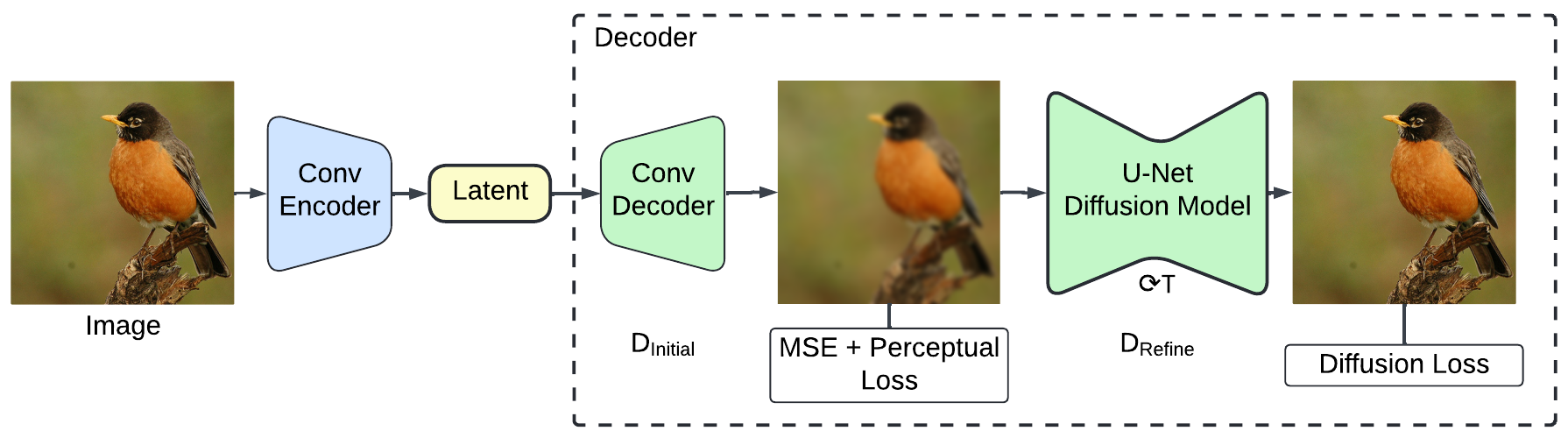}
    \caption{
    A block diagram of our auto-encoder architecture.
    Our "Diffusion Model" is a U-Net as defined by \cite{simple_diffusion}.
    During inference, the diffusion model is run in a loop
    to iteratively sample an image. All of these parameters jointly after
    being initialized from scratch, except for the weights of the network
    used in the perceptual loss.}
    \label{fig:diagram}
\end{figure}

\begin{figure}[h]
    \centering
    \begin{subfigure}{0.35\textwidth}
      \includegraphics[width=\textwidth]{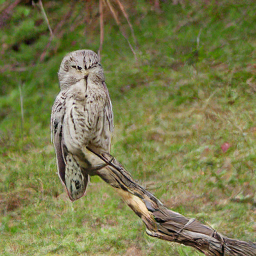}
    \end{subfigure}~
    \begin{subfigure}{0.35\textwidth}
      \includegraphics[width=\textwidth]{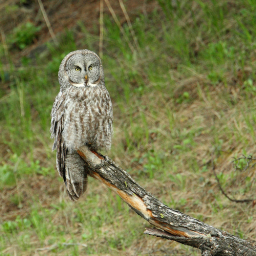}
    \end{subfigure}
    
    \begin{subfigure}{0.35\textwidth}
      \includegraphics[width=\textwidth]{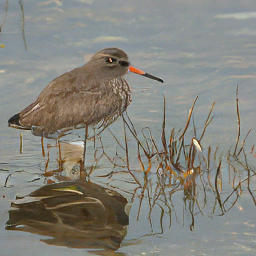}
    \end{subfigure}~
    \begin{subfigure}{0.35\textwidth}
      \includegraphics[width=\textwidth]{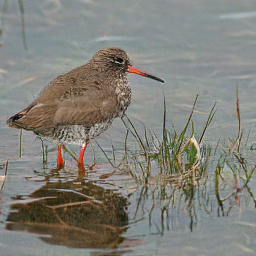}
    \end{subfigure}
    

    \begin{subfigure}{0.35\textwidth}
      \includegraphics[width=\textwidth]{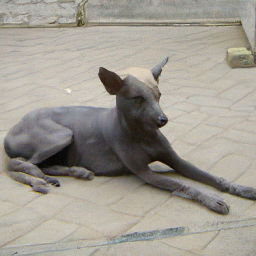}
      \caption{No auxiliary losses.}
    \end{subfigure}~
    \begin{subfigure}{0.35\textwidth}
      \includegraphics[width=\textwidth]{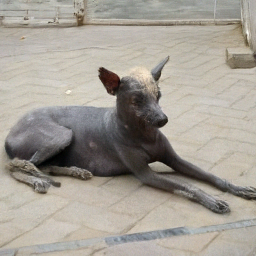}
      \caption{Only perceptual auxiliary loss.}

    \end{subfigure}
    \caption{Visual impact of perceptual loss.}
    \label{fig:perceptual_images}
\end{figure}

We find that additional direct penalization of $\di(E(x)$ leads to improved CMMD and FID and less distortion (see Figure~\ref{fig:perceptual_images}). This was achieved by minimizing a composite loss
containing terms with favorable Hessian (\cref{eq:mse}) and perceptual characteristics (\cref{eq:perceptual}),
\begin{align}
\label{eq:total}
\ell_{\mathrm{Total}} \defeq \ell_{\text{AE}} + \lambda_{p}\ell_{\mathrm{Perceptual}}
+ \lambda_m\ell_{\mathrm{MSE}}
\end{align}
where,
\begin{align}
\label{eq:mse}
\ell_{\mathrm{MSE}} \defeq \left\| x - \di(E(x))\right\|_2^2
\end{align}
and,
\begin{align}
\label{eq:perceptual}
\ell_{\mathrm{Perceptual}} \defeq \left\|f_\mathrm{Frozen}(x) - f_\mathrm{Frozen}\Big( \di(E(x))\Big)\right\|_2^2.
\end{align}
The function $f_\mathrm{Frozen}$ is an unlearnable standard ResNet, itself trained on 
ImageNet and used for both the baseline and \algoname. We found the best hyper-parameter setting for \cref{eq:total} is $\lambda_m = 1$ and $\lambda_p = 0.1$
Setting $\lambda_p > 0$ was particularly important to be competitive at reconstruction with GAN based methods in Table~\ref{tab:losses}. 
The visual impact of perceptual loss is shown in Figure~\ref{fig:perceptual_images}. 

For generating reconstructions (recall that the \algoname~decoder is stochastic) we used classifier-free guidance during inference \citep{cfg}; for the unconditional model the U-Net was trained with $\di(E(x))$ dropped-out, i.e., randomly zeroed out on 10\% of training instances.

\section{Experiments}\label{sec:experiments}

\label{sec:compression}
In this section we explore how the GAN-based loss compares to our approach.
Without loss of generality, we define the relative compression
ratio of $1$ to be a network that maps $ 8 \times 8$ RGB patches to an $8$ dimensional
latent vector. Effectively, this means for our encoder 
$E$ if and $x\in \reals^{256 \times 256 \times 3}$,
then $E(x) \in \reals^{32 \times 32 \times C}$
where $C=8$.
In general, to achieve a relative compression ratio of 
$K$ we set $C = \frac{8}{K}$. The effect of increasing
the compression ratio is plotted in Figure~\ref{fig:compression}.
Observe that distortion degrades much more rapidly for the
GAN based auto-encoder as measured by CMMD \citep{cmmd} which
Imagen-3 \citep{imagen3} showed better correlates with
human perception.

Not only is our approach better at all compression levels; the gap between the GAN based
autoencoder and \algoname{} widens as we increase the relative compression ratio.
Using the much simpler diffusion formulation under Equation~\ref{eq:total},
we are able to reconstruct crisp looking images with detailed textures
(See Figure~\ref{fig:pictures}). Our method has the added benefit that
we do not need to tune any GAN related hyper-parameters, and can scale up effectively 
using the large body of diffusion literature (\citet{edm}, \citet{diffusionguide}).

\subsection{Impact of $\di$}
\label{sec:di}
\begin{figure}
    \begin{minipage}{.49\textwidth}
    \centering
    \begin{tabular}{llr}
        \toprule
        $\lambda_p$ & $\lambda_m$ & CMMD\ $\downarrow$ \\
        \midrule
        0 & 0 & 0.43 \\
        0 & 1 & 0.32 \\
        0.1 & 0 & 0.13 \\
        0.1 & 1 & 0.15\\
        \bottomrule
    \end{tabular}
    \captionof{table}{Individual impact of each of our auxiliary losses. We note
    that the perceptual loss has a big impact, and is crucial to
    being competitive while training autoencoders. See image samples
    in Figure~\ref{fig:perceptual_images}.
    The first row $\lambda_p = \lambda_m = 0$ is analogous
    to $\di$ having no role to play. 
    }
    \label{tab:losses}
    \ \\ 
    \ \\
    \ \\ 
    \end{minipage}
    \begin{minipage}{0.49\textwidth}
        \centering
        \includegraphics[width=0.88\textwidth]{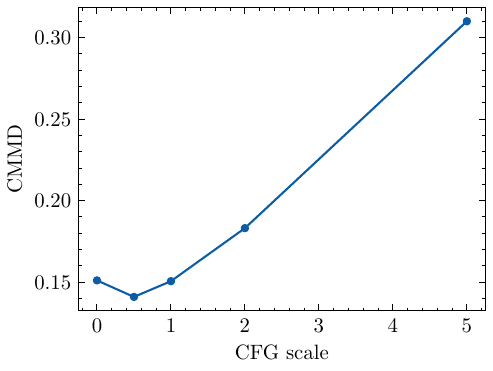}
        \captionof{figure}{Effect of CFG scale in $\dr$.We use CFG scale of $0.5$ in our experiments.}
        \label{fig:cfg}
    \end{minipage}
\end{figure}
Observing Equation~\ref{eq:aeloss} and Figure~\ref{fig:diagram}, we note
that the output of $\di$ is an intermediate tensor that is not strictly
required for the diffusion loss or for generating the output. We
show using Table~\ref{tab:losses} that this piece is crucial for
achieving performance comparable to the GAN based autoencoder.
The perceptual loss term in particular has a large impact in
reducing distortion. Visual examples are shown in Figure~\ref{fig:perceptual_images}. 
We found that separating the decoder into 2 parts was necessary.
When we applied  LPIPS and MSE losses without
$\di$, our method was not competitive with GAN based
autoencoders.

\subsection{Analysis of sampling in $\dr$}
\label{sec:dr}
\begin{figure}
\centering
\begin{subfigure}{0.15\textwidth}
\captionsetup{labelformat=empty}
\includegraphics[width=\textwidth]{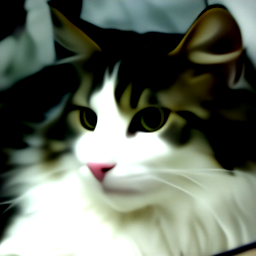}
\caption{2 steps}
\end{subfigure}
\begin{subfigure}{0.15\textwidth}
\captionsetup{labelformat=empty}
\includegraphics[width=\textwidth]{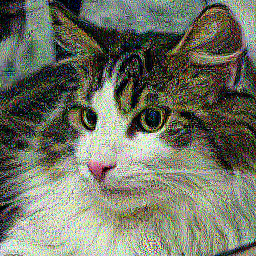}
\caption{5 steps}
\end{subfigure}
\begin{subfigure}{0.15\textwidth}
\captionsetup{labelformat=empty}
\includegraphics[width=\textwidth]{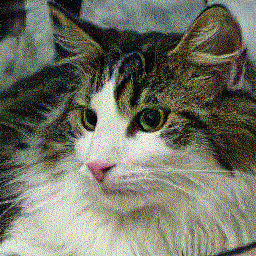}
\caption{10 steps}
\end{subfigure}
\begin{subfigure}{0.15\textwidth}
\captionsetup{labelformat=empty}
\includegraphics[width=\textwidth]{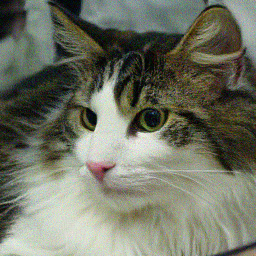}
\caption{50 steps}
\end{subfigure}
\begin{subfigure}{0.15\textwidth}
\captionsetup{labelformat=empty}
\includegraphics[width=\textwidth]{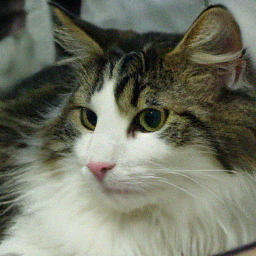}
\caption{150 steps}
\end{subfigure}
\caption{Example of how reconstructions look when changing the number of
sampling steps for $\dr$. We can see
that even with steps $\le5$, the high level structure of the image is preserved.
When our method is used for interactive generation (for example, with a latent diffusion model), we can use fewer steps of $\dr$ to show the user multiple low-quality inputs,
and use a high number of steps for the final generation that the user selects.}
\label{fig:catsteps}
\end{figure}

\begin{figure}
\centering
\captionbox{Impact of number of sampling steps on $\dr$ on CMMD. \label{fig:cmmdsteps} }
{\includegraphics[width=0.45\textwidth]{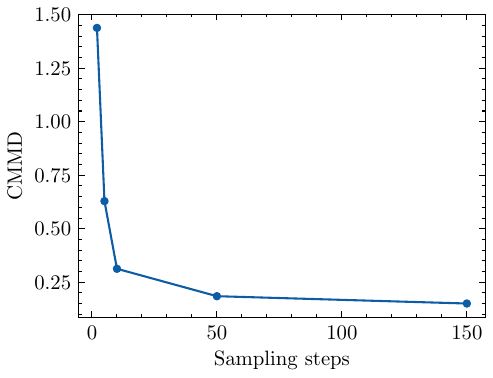}}
\hfill
\captionbox{\label{fig:psnr}
Rate-distortion-perception trade-off as outlined by \citet{rethink_lossy_compression}. Despite higher distortion, \algoname{} is perceptually better.}
{\includegraphics[width=0.44\textwidth]{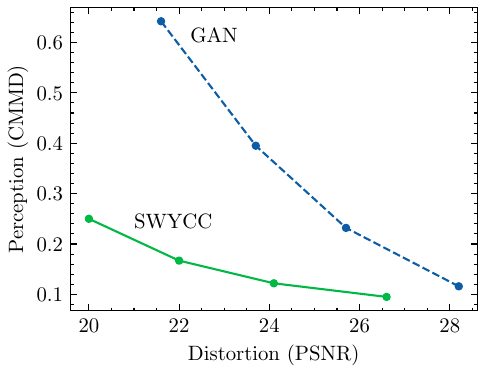}}
\end{figure}

In Figure~\ref{fig:catsteps} we show the qualitative difference number of
sampling steps makes to reconstructions. We can see that even with just
$2$ steps the high level structure of the image is present. In Figure~\ref{fig:cmmdsteps}
we study the impact of number of sampling steps by using the CMMD metric.
Figure~\ref{fig:samplediff} shows what sampling in $\dr$ actually ends up changing.
We can see that only regions with high-frequency components and detailed textures
are changed between samples, while regions containing similar colors over large
areas are left untouched.

Figure~\ref{fig:cfg} studies the effect of classifier-free guidance \citep{cfg}
as used in $\dr$. We ablate the guidance with a model trained at a relative compression
factor of $4$ (See Section~\ref{sec:compression} for definition) and find
that a guidance scale of $0.5$ works the best. This is not to be
confused with the guidance scale of the latent diffusion model that may
be trained on top of our autoencoder, which is a completely separate parameter
to be tuned independently. 

\subsection{Modeling latents for diffusion}
\label{sec:ldm}
We use a DiT model \citep{dit} to model the latent space of our models
for the task of class-conditional image generation.
In Figure~\ref{fig:generation} we compare our latents with those of a GAN based autoencoder.
Our approach leads to $5\%$ lower FID \citep{fid} than the GAN baseline at the best $4\times$ compression ratio.
Notably our approach achieves its best result at the highest compression ratio, where the task of modelling the latent representation is simplest, whereas the GAN autoencoder is unable
to operate effectively in this regime.


\subsection{Exploring better perceptual losses}
\begin{figure}
\captionbox{\label{fig:samplediff}
Reconstructed images and a heat-map of variance between 10 samples.}
{
\centering
\begin{subfigure}{0.15\textwidth}
\includegraphics[width=1\textwidth]{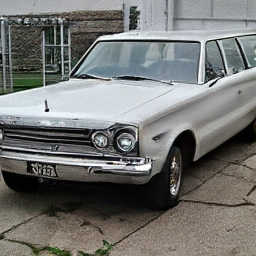}\\
\includegraphics[width=1\textwidth]{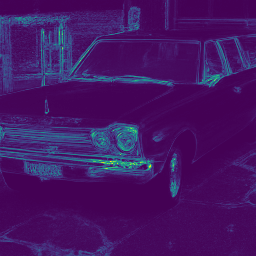}
\end{subfigure}
\begin{subfigure}{0.15\textwidth}
\includegraphics[width=1\textwidth]{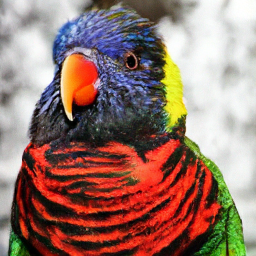}\\
\includegraphics[width=1\textwidth]{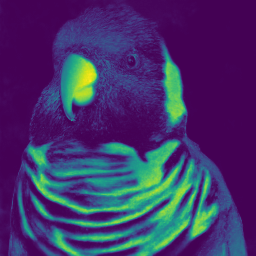}
\end{subfigure}
\begin{subfigure}{0.15\textwidth}
\includegraphics[width=1\textwidth]{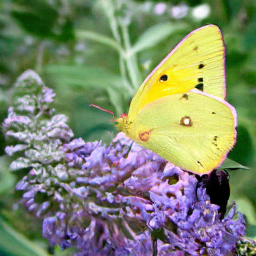}\\
\includegraphics[width=1\textwidth]{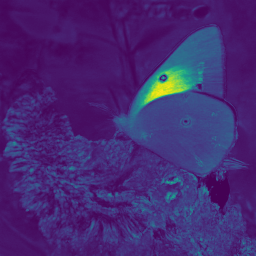}
\end{subfigure}
\vspace{12pt}
}
\hfill
\captionbox{\label{fig:dists}
Effect of using DISTS loss on reconstruction quality.}
{\includegraphics[width=0.45\textwidth]{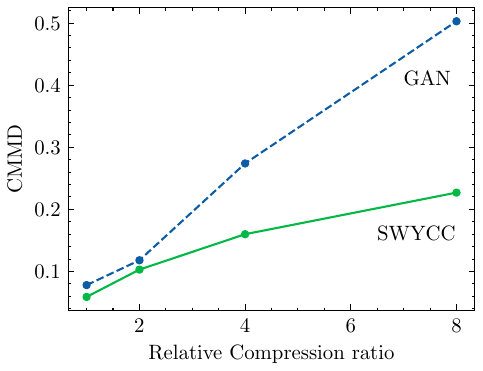}
\vspace{-5pt}}
\end{figure}

In Table~\ref{tab:losses} we showed the large impact perceptual loss
has on reconstruction quality. This begs the question; are there
better auxiliary losses we can use?
We compare the perceptual loss as described by in VQGAN~\citep{vqgan, perceptual_style_transfer} and replace it with the DISTS~\citep{dists}.
DISTS loss differs from perceptual loss in 2 important ways. a)
It uses SSIM~\citep{ssim} instead of mean squared error and b) It
uses features are multiple levels instead of using only
the activation's from the last layer.
The results are shown in Figure~\ref{fig:dists}.
At lower relative compression ratios, DISTS loss helps GANs
and \algoname{}. But at higher relative compression ratios,
GAN based autoencoders perform even worse than the perceptual
loss based baseline. We think this is an avenue for future exploration. 
We perform all other experiments with perceptual loss~\citep{vqgan},
since it is more prevalent in literature and it helps
GAN based auto-encoders at higher relative compression ratios.

\subsection{Architecture and hyper-parameters}
{\bf Autoencoder training hyper-parameters}
We train all of our models on the ImageNet dataset resized at $256 \times 256$
resolution. During training, we resize the image such that the shorter side measures
$256$ pixels and take a random crop in that image of size $256 \times 256$. For measuring reference statistics on the validation set, we take the largest
possible center square crop. All of our models are trained at a batch size of $256$ for
$10^6$ steps which roughly equals $200$ epochs. 

{\bf GAN baseline:}
We use the popular convolutional encoder-decoder architecture popularized
by MaskGIT \citep{maskgit} in our GAN-based baselines. This autoencoder design is used by many image models, including FSQ \citep{fsq} and GIVT \citep{givt}, and was extended to the video domain by MAGVIT-v2 \citep{magvitv2}, VideoPoet \citep{kondratyuk2024videopoet}, and WALT \citep{walt}. We take advantage of decoder improvements developed by MAGVIT-v2 (see Section 3.2 in \citep{magvitv2}) to improve reconstruction quality.
Note that while the video models enhance the base autoencoder architecture with 3D convolution to integrate information across time, the discriminator and perceptual loss are still applied on a per-frame basis and thus are essentially unchanged in our model.

{\bf \algoname{}:} In our experiments we keep the architecture of $\di$ identical
to the decoder used in the GAN baseline. For $\dr$ we use the U-Net architecture
as parameterized by \cite{simple_diffusion}. We borrow the U-Net 256 architecture
and make the following modifications:
\texttt{\begin{itemize}[label={},topsep=0pt,itemsep=-1ex,partopsep=1ex,parsep=1ex]
\item channel\_multiplier = [1, 2, 4, 8]
\item num\_res\_blocks = [2, 4, 8, 8]
\item downsampling\_factor = [1, 2, 2, 2]
\item attn\_resolutions = [16]
\item dropout = 0.0
\end{itemize}
}
We train using $v$-parameterization \citep{progdistill}, which corresponds to $w_t=\sigma_t^{-2}$ in \eqref{eq:diffusion-loss}.
We use the Adam optimizer to learn our parameters. The learning rate is warmed up
for $10^4$ steps from $0$ to a maximum value of $10^{-4}$ and cosine decayed
to $0$. We use gradient clipping with global norm set to $1$.

{\bf Latent Diffusion}:
All of our experiments are done with the DiT-L architecture with $2\times2$ patching \citep{dit} with the addition of SwiGLU \citep{Shazeer2020GLUVI} and 2D RoPE \citep{heo2024rotarypositionembeddingvision}.
We train for $ 4 \times 10^5$ steps with a batch size of $256$, using a constant learning rate of $10^{-4}$ and dropout the class embedding $10\%$ of the time during training.
For inference we use a classifier-free guidance scale of $0.5$
which gave optimal results in \cite{dit}.

\section{Related Work}\label{sec:related}
{\bf Autoencoders for 2-stage generation: }For discrete representation learning, \citet{vqvae} showed the usefulness of the 2-stage modeling approach. In this broad category, the first stage fits an autoencoder to the training data with the goal of learning a compressed representation useful for reconstructing images. This is followed by a second stage where the encoder is frozen and a generative model is trained to predict the latent representation based on a conditioning signal. This approach regained popularity when \citet{dalle} showed that it can be used for zero-shot 
text generation, and is now the dominant approach for image and video generation \citep{maskgit,magvitv2,chang2023muse,walt,kondratyuk2024videopoet}.

{\bf Adversarial losses: }\citet{vqgan} extended the autoencoder from \citet{vqvae} with
two important new losses, the perceptual loss and the adversarial loss,
taking inspiration from the works of \citet{perceptual_style_transfer}
and \cite{image2image_patchdisc}. The perceptual loss is usually defined as the L2 loss between a latent representation of the original and reconstructed image.
The latent representation, for example, can be extract from
the final layer activations of a ResNet optimized to classify ImageNet images.
The adversarial loss is a patch-based discriminator that uses a discriminator
network to predict at a patch-level whether it is real or fake.
This encourages the decoder to produce realistic looking textures.

{\bf Latent Diffusion: } \citet{stable_diffusion} popularized text-to-image
generation using latent diffusion models. They kept the autoencoder
from \citet{vqgan} intact and simply removed the quantization layer.
This accelerated diffusion model research in the community owing
to the fact that the latent space was much smaller than the
pixel space, which allows fast training and inference compared
to diffusion models like Imagen that sample pixels directly \citep{imagen}.

{\bf 2-stage diffusion autoencoders:}
\citet{diffusevae} and \citet{diffusionae} both 
train decoders with a diffusion loss. The crucial
difference is that both these works train their
autoencoder in 2-stages. 

{\bf Compression and diffusion:} \citet{diffusion_compression_luca} 
showed that diffusion models can be used for compression. Crucially,
compared to our approach they use a frozen autoencoder, and do not
train their autoencoder end-to-end. They also use an objective based on modified flow matching. In contrast, we did not modify the loss or the sampling algorithm. \citet{diffusevae} use a similar
approach with a 2-stage autoencoder training process for their autoencoder.

In a similar context, \citet{yang2024neurips} developed an end-to-end optimized compression model using a diffusion decoder. They show improved perceptual quality compared to earlier GAN-based compression methods at the expense of higher distortion (pixel-level reconstruction accuracy). Different from our approach, they use a discrete latent space, which is required for state-of-the-art compression rates achieved via entropy coding. This limits the reconstruction quality but is required for a compression model that ultimately seeks to minimize a rate-distortion objective, not just a reconstruction and sampling quality objective.

Würstchen architecture \citep{stable_cascade} has shown that training a cascade of diffusion models
improves training efficiency. Crucially Würstchen, does
not apply a diffusion loss on pixels, instead resorting to 
a GAN loss.
\citet{deblur} also
pointed out that diffusion can fix a lot of pixel
level artifacts, although they do not investigate
training autoencoders.

\cite{divae} learn an autoencoder using a diffusion
loss for discrete vector quantized encodings.
We differ from them in 2 crucial ways; (i) 
we learn a continuous representation 
(ii) We show that our architecture produces
latents that are better for latent diffision
by showing that we can achieve 
\emph{higher compression and better generation performance
(Figure~\ref{fig:generation})
}.

\section{Conclusion}\label{sec:conclusion}
We have described a general autoencoder framework that uses a diffusion based
decoder. Compared to decoders that use GANs, our system is much more easier
to tune and has the same theoretical underpinnings as diffusion models.
We showed our method produces sigificantly less distortions as compared
to GAN based autoencoders in Figure~\ref{fig:compression} and are better
behaved as latent spaces for diffusion in Figure~\ref{fig:generation}.
In Section~\ref{sec:di} and~\ref{sec:dr} we studied
the hyper-parameter settings on the 2 major components of our decoder,
$\di$ and $\dr$.

{\bf Possible extensions:} The autoencoder technique we describe is fairly
general and can be extended to any other continuous modality like audio, video or 
3D-point clouds. In addition, all improvements to diffusion algorithms like those
by \citet{diffusionguide} can be carried over.

{\bf Limitations:} The main limitation of our method is the increase in inference
cost during decoding. This can be partly mitigated by using fewer steps like in Figure~\ref{fig:catsteps}. In addition, techniques used to improve diffusion
sampling time like Progressive distillation \citep{progdistill} and Instaflow \citep{instaflow}
are also prudent. 
Because of $\dr$, our training time compute cost
is also higher. Combining $\di$ and $\dr$ in a clever
way to reduce training time compute and memory could be a promising research direction.

{\bf Acknowledgements: }
We would like to thank Thomas Mensink
for helping us find better 
U-Net and diffusion baselines. We also
 thank Saurabh Saxena for his help
 with navigating an internal
 diffusion codebase.

\bibliography{iclr2025_conference}
\bibliographystyle{iclr2025_conference}

\newpage
\section{Appendix}\label{sec:appendix}
\subsection{Additional metrics}
\begin{table}[ht]
\centering
\begin{tabular}{lrrrrrr}
\toprule
Method & Rel. Compression & rFID & Inception Score & pSNR & CMMD & rFID(Dino-v2) \\
\midrule
\algoname{} & 1 & 0.33 & 222 & 26.6 & 0.095 & 4.13 \\
 & 2 & 0.62 & 215 & 24.1 & 0.122 & 9.56 \\
 & 4 & 1.17 & 203 & 22.0 & 0.167 & 22.4 \\
 & 8 & 2.75 & 175 & 20.0 & 0.250 & 60.1 \\
GAN & 1 & 0.27 & 222 & 28.2 & 0.116 & 4.52 \\
 & 2 & 0.54 & 215 & 25.7 & 0.232 & 11.7 \\
 & 4 & 0.99 & 202 & 23.7 & 0.395 & 30.3 \\
 & 8 & 1.96 & 176 & 21.6 & 0.642 & 72.9 \\
 SD-VAE 2.x (on COCO) &  2 & 4.70 & & 24.5 \\ 
\bottomrule
\end{tabular}
\caption{Additional reconstruction metrics on ImageNet, unless otherwise noted.  \algoname{}
is better than GAN at all perceptual metrics
except rFID.}
\label{tab:allmetrics}
\end{table}

\subsection{Parameter counts}
\begin{table}[ht]
\centering
\begin{tabular}{llr}
\toprule
Method & Component & Parameters (Million)
 \\
\midrule
\algoname{} & Encoder & 49.4\\
& $\di$ & 63.4\\
& $\dr$ & 614.1\\
GAN & Encoder & 49.4\\
 & Decoder & 63.4\\
\bottomrule
\end{tabular}
\caption{Parameter counts of network components.}
\label{tab:paramcount}
\end{table}

\newpage
\subsection{Visual examples}


\end{document}